\documentclass{article}
\usepackage{ijcai23}

\usepackage{times}
\usepackage{soul}
\usepackage{url}
\usepackage[hidelinks]{hyperref}
\usepackage[utf8]{inputenc}
\usepackage[small]{caption}
\usepackage{graphicx}
\usepackage{amsmath}
\usepackage{amsthm}
\usepackage{booktabs}
\usepackage{algorithm}
\usepackage{algorithmic}
\usepackage[switch]{lineno}
\usepackage[font={small}, skip={18pt}]{caption}
\usepackage{dblfloatfix}
\usepackage{tabularx}
\usepackage{float}
\usepackage{amsmath,amssymb}

\title{Deep Generative Multi-Agent Imitation Model as a Computational Benchmark for Evaluating Human Performance in Complex Interactive Tasks: A Case Study in Football}
\author{
    Chaoyi Gu, Varuna De Silva
}
\begin{document}
\maketitle
\begin{abstract}
    Evaluating the performance of human is a common need across many applications, such as in engineering and sports. When evaluating human performance in completing complex and interactive tasks, the most common way is to use a metric having been proved efficient for that context, or to use subjective measurement techniques. However, this can be an error prone and unreliable process since static metrics cannot capture all the complex contexts associated with such tasks and biases exist in subjective measurement. The objective of our research is to create data-driven AI agents as computational benchmarks to evaluate human performance in solving difficult tasks involving multiple humans and contextual factors. We demonstrate this within the context of football performance analysis. We train a generative model based on Conditional Variational Recurrent Neural Network (VRNN) Model on a large player and ball tracking dataset. The trained model is used to imitate the interactions between two teams and predict the performance from each team. Then the trained Conditional VRNN Model is used as a benchmark to evaluate team performance. The experimental results on Premier League football dataset demonstrates the usefulness of our method to existing state-of-the-art static metric used in football analytics.
\end{abstract}

\section{Introduction}
Evaluation of human performance in solving complex tasks is needed in lots of areas, such as in sports and engineering. The most widely used method is to use a metric which has been proved to be valid and effective for specific context.However, this method can still be unreliable when the existing metric cannot capture the complicated contexts. For example in sports performance analysis, how do you measure the quality of attacking or defence of footballer/ basketball player, playing under different levels of pressure? In these occasions, the result of evaluation is often based on the coaching staff's expertise which can unavoidably be subjective and biased. In this study, we are interested in building a model which can be used as benchmark to evaluate human performance in solving difficult tasks involving multiple humans and complicated contextual factors. We demonstrate this in the context of football team and player performance analysis. In each possession of one football game, the team needs to accomplish a series of complex tasks to reach the best outcome. In attacking, the players need to pass or dribble to create chances for scoring the goal, while in defending, the players are required to win the ball back or protect the area near the goal preventing opponent from making shots. In both kinds of scenarios, not just individual performance but the team cooperation as well, will make a difference in deciding the outcome.

\graphicspath{ {./images/} }
\begin{figure}
    \centering
    \includegraphics[width=2.65in]{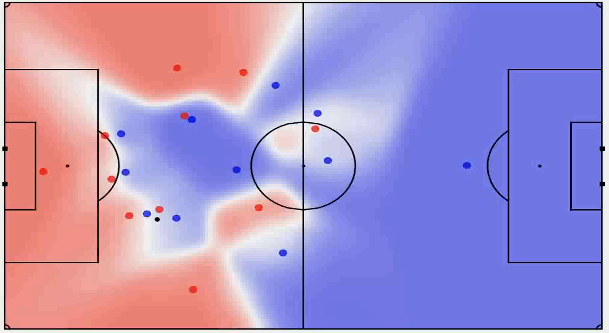}
    \caption{Illustration of a Pitch Control Map}
    \label{fig:pcm}
\end{figure}

In recent studies, the researcher has built a Xg model (goal probability model) based on synchronised positional and event data to predict the goal probability of given game state and evaluate player and team performance according to their contribution to the increase of goal probability \cite{anzer2021goal}. Other research specifically looked into the passing within football game, proposing a framework to assess passing effectiveness in professional football games based on the number of defenders getting passed after the pass and the difference of space controlled by the attacking team before and after the pass \cite{rein2017pass}. Both above research looked partially into football game when making evaluation of players, at either the shooting or the passing without involving more complicated contextual factors when building their models. There are state-of-the art metrics making a more thorough evaluation of player performance taking more contextual factors into consideration, one of which is VAEP for valuing player on-ball actions in football \cite{decroos2019actions}. They put contextual information such as the trajectories of players and ball before the specific game event into independent variables and the outcome (scoring or conceding a goal) after the event as the dependent variable to build the event evaluation model and use the model to generate a metric for evaluating each player's performance which is related to specific game event such as passing or dribbling. The limitation of the VAEP is that the model results heavily rely on the patterns learned from training data, which causes the potential subjectivity and inaccuracy when analysing player performance in new contexts considering the complexity of football game. The other state-of-the-art metric for evaluating human performance in playing football is the EPV(Expected Possession Value) \cite{fernandez2019decomposing}. For a given game state in the football game, they model the potential value in terms of contribution to scoring from passing, dribbling and shooting. The limitation of EPV is that when quantifying the expected value from certain events such as passing to specific location of the pitch, they didn't consider the interaction between two teams and among individual players. Specifically, when the pass is made, the opponent team will make a reaction in terms of team movement instead of standing still, which will change the contextual factors. Moreover, Imitation learning, a form of supervised machine learning training agents to act like human by demonstrating the desired behaviours to them \cite{hussein2017imitation}, is used in analysing human behaviors in completing difficult tasks such as driving and playing sports games. It is used in football performance analysis to evaluate player performance in defending comparing it against performance from benchmark AI agents by training a group of AI defenders to act like human players through learning from mapping between observations and actions \cite{le2017coordinated}. The limitation of Imitation Learning is that obtaining expert demonstrations is expensive in applications, for instance in football it is hard to map game-play data to expert demonstrations for attacking which consists of complex contextual information. The State-of-the-art Imitation Learning model in football performance analysis can only set the benchmark for defending when evaluating team performance and it is unable to capture the interactions between two opposite teams making it unreliable when used to do analysis.

In this work, we build a deep generative multi-agent imitation model to deal with the aforementioned challenges. We firstly map the team movements in each possession within football games to a sequence of Pitch Control Maps which computes and visualizes the space controlled by each team as shown in Figure \ref{fig:pcm} \cite{spearman2017physics}. These Pitch Control Maps include not only geometric contextual information of the game but also the information about interactions between two opposite teams and cooperation among players within the same team. Then we model each sequence of game-play utilizing Conditional Variational Recurrent Networks to imitate the players' movements and interactions. We use the trained model to predict the sequence of play with the first frame of sequence as the input. Then the predicted game-play is used as a benchmark for comparing the real team performance against and evaluate performance from team and individual levels. Experimental results show that the proposed model benchmark is a good addition to existing state-of-the-art metrics in evaluating human performance in playing football.

The contributions of our research can be summarized as follows:
\begin{itemize}
\item We propose a novel approach to evaluate football players' performance by building a benchmark by deep generative modelling.

\item The results of our model on a Premier League dataset shows our model is efficient in performance evaluation in football and it is a good addition of existing evaluating metrics.

\item The effectiveness of applying our model in football shows the potential of utilizing similar framework to evaluate human performance solving other complex tasks.
\end{itemize}

\section{Background}

\subsection{Football Performance Analysis}
\subsubsection{Pitch Control Map}
A Pitch Control Map computes the area controlled by each team within a given game state of the football game \cite{spearman2017physics}. As shown in Figure \ref{fig:pcm}, the red area denotes the area controlled by the attacking team and the blue area represents the the area controlled by the defensive side. At first, a passing probability model is trained to predict the successful rate for a pass given positions, speed and facing directions of ball and players, and the player ids of passer and receiver. As shown in Eq. \eqref{eq:1}, every pass is viewed as a Bernoulli trial with k representing the outcome of the pass (success: 1, failure: 0). The probability that the pass will be successful is derived by input x and two trained parameters \(\sigma\) and \(\lambda\). 
\begin{equation}
        \label{eq:1}
        p\left(k\middle|\sigma,\ \lambda,x\right)=
        \begin{cases}
            1-p\ for\ k=0\\
            p\ \ \ \ \ \ \ \ for\ k=1
        \end{cases}
\end{equation}%
The pass probability model is then trained to find a set of parameters maximizing the likelihood shown in Eq. \eqref{eq:2}. 
\begin{equation}
    \label{eq:2}
    \mathcal{L}\left(\sigma,\ \lambda\middle|\ k,x\right)=\ P(k|\sigma,\ \lambda,x)
\end{equation}%
Secondly, the trained pass probability model is used to compute the stationary pass probability for an imagined ball location and convert the whole pitch into a real valued scalar field for generating a Pitch Control Map.
Pitch Control Maps have been used by coaching staff to analyse football game-plays since it indicates the effect of players' movements and interactions on team control of the space.

\subsubsection{Expected Possession Value (EPV)}
EPV in football was first introduced in 2011 \cite{rudd_2011}. It quantifies the ball possession value given the ball position and match state such as open play or set piece (Eq. \eqref{eq:3}). Each possession within the football game is defined as a sequence of game states with the possible next state only depending on the current state, following the theorem of Markov Process \cite{stroock2013introduction}.

Game states include zones denoting the ball's location and the outcome of the possession (scoring or losing possession). Unlike Xg model calculating the possibility of scoring when shooting at given location \cite{rathke2017examination}, EPV model computes the probability of current possession developing into a goal given ball location. 
\begin{equation}
    \label{eq:3}
    P_{poss}\left(G\middle|situation\right)=\ P_{poss}(G|ball,\ match\ state)
\end{equation}%
After the release of original EPV model, several advanced models were introduced on top of that with more contextual factors such as positions of players involved. The state-of-the-art EPV model was introduced in 2019 \cite{fernandez2019decomposing}. The model is composed by three independent sub-models for passing, shooting and ball-driving. It computes the expected value of current possession based on all the spatiotemporal information related to players and ball.
In this work, we use the original version of the EPV model \cite{rudd_2011} when quantifying our proposed team performance benchmark since the benchmark is represented by images which don't contain enough contextual information for running the advanced version of EPV model \cite{fernandez2019decomposing}.
\subsection{Deep Generative Model}
\subsubsection{Variational Autoencoder}
A Variational Autoencoder (VAE) generates new samples by learning input data's hidden representation in terms of regularities and patterns \cite{kingma2013auto}.The main idea of VAE is to extract low dimensional latent space from key features of inputs, such as images, by encoding. Compared with autoencoder having similar architectures with encoder and decoder, VAE can overcome overfitting in most contexts especially those with complicated input data in order to generate more reasonable and consistent new samples. VAE introduces a generative process with prior \(p(z)\) and complex likelihood \(p(x|z)\). \(z\) denotes a set of random variables and x denotes observed variables. VAE uses variational neural network \(q(z|x)\) enabling the utilization of lower bound, to approximate the intractable posterior \(p(z|x)\):
\begin{equation}
    logp(x)\geq-KL(q(z|x)||p(z)) + \mathbb{E}_{q(z|x)}[logp(x|z)]
\end{equation}
in which \(KL(Q||P)\) represents Kullback-Leibler divergence between distribution \(Q\) and distribution \(P\). \(q(z|x)\), as the approximate posterior is a Gaussian distribution \(N(\mu, \sigma^2)\) where \(\mu\) as mean and \(\sigma^2\) as variance are the output of a non-linear function of \(x\) as a neural network. Through maximizing the variational lower bound, the generative model \(p(x|z)\) and inference model \(q(z|x)\) are trained jointly. In the variational lower bound, the integral related to  \(q(z|x)\) is approximated stochastically; the gradient of this estimated integral is enabled to have a low variance by utilizing reparameterization trick. \(z\) is reparameterized to \(\mu+\sigma\bigodot\epsilon\) in which \(\epsilon\) denotes a vector of standard Gaussian distribution variables. Besides, to train the inference model, standard backpropagation for stochastic gradient descent is used.
\subsubsection{Variational Recurrent Neural Network}
Variational Recurrent Neural Network (VRNN) is an extension of VAE combining VAE by utilizing recurrent neural networks to model sequential data. It models the reliance between random variables through subsequent timesteps. There is a VAE model at every timestep of VRNN while parameters of each VAE are conditioned on the recurrent neural network's state variable. This structure allows VRNN to capture the temporal structure of sequential data when being trained so that it can be used to generate new sequential data samples. Since the data used in this work are sequences of game-play with complex contextual information, we conditioning the model inference and generation on the labels showing the pattern of players movements and interaction to enable a more consistent generation of benchmark team performance. We propose a new architecture on top of the Vanilla VRNN model to achieve this objective.
\section{Methodology}
To reach the research objective which is to evaluate football players' performance based on benchmarks set by computational AI agents, we build the benchmark in following steps. Firstly we map the existing player performance data into sequences of Pitch Control Maps which demonstrate the cooperation between teammates and interactions between rivals, in terms of space-control of the pitch. Then we train a generative model on Pitch Control Map sequences to generate benchmark AI performance in a new given football context.
\subsection{Dataset}
Tracking and event data of 43 matches from Premier League are used in this work. Tracking and event data are synchronized to form a comprehensive dataset recording contextual information related to every possession of game-play from these matches. Besides, we remove all the set-pieces, penalties and events when game is interrupted with only active plays remained.
\subsection{Data Pre-Processing}
We use tracking and event data to generate Pitch Control Maps for each possession of active game plays. Each possession is converted to a sequence of Pitch Control Maps with length varying from 2 to 110. The time interval between each two adjacent Pitch Control Maps is 1 second. There are 200,000 sequences in total from our dataset.
10 per cent of the dataset, 20,000 sequences are used to train CNN-LSTM Classifier. To meet the requirement of input shape for training the classifier, sequences including more than 2 frames are processed by a sliding window to generate the input sequence with the same length as 2. All the sequences are manually labelled as pushing, backing or unlabelled. There are 65,000 sequences in total after above processing, in which we use 80 per cent for training and the rest for testing. 
The other 90 per cent of the dataset, including 180,000 sequences of Pitch Control Maps are used to train Conditional VRNN Model. We set the length of input sequence as 6. We firstly filtered sequences having less than 6 frames, with 125,000 sequences left. Secondly the sequences are processed by a sliding window to generate the sequences with same length as 6. There are 500,000 sequences remained after processing, which are split into training and testing dataset for Conditional VRNN model with 400,000 and 100,000 samples included respectively.

\subsection{Pre-trained Classifier}
\begin{figure}
    \centering
    \includegraphics{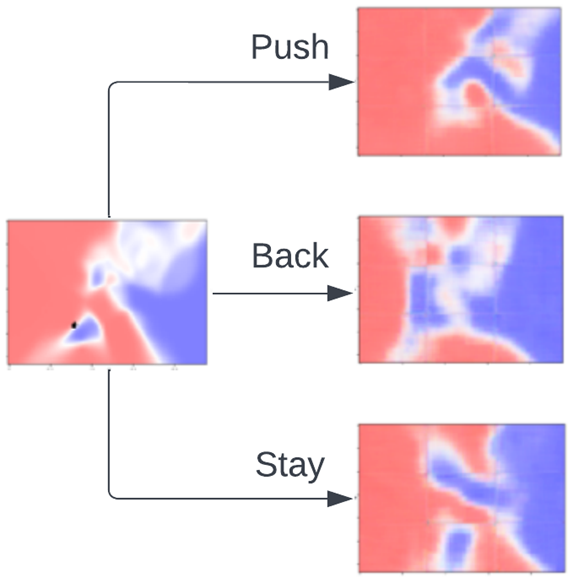}
    \caption{Three types of Pitch Control Map Patterns}
    \label{fig:mp}
\end{figure}

To label the sample data for training Conditional VRNN model, we firstly train a classifier using CNN-LSTM architecture to classify the Pitch Control Maps according to its dependence on the frame from previous timestep. As shown in Figure \ref{fig:mp}, there are three types of patterns in Pitch Control Map changes, pushing, backing and staying regarding the change of controlled area by attacking side.
We label the sequences showing pushing or backing patterns while the sequences showing unnoticeable pattern changes are left unlabelled. We train the classifier using labelled data while in prediction the sequences with confidence less than 0.95 are believed to be those showing no detectable patterns of pushing or backing and these sequences are classified as staying.

\begin{figure}
    \centering
    \includegraphics[width=\linewidth]{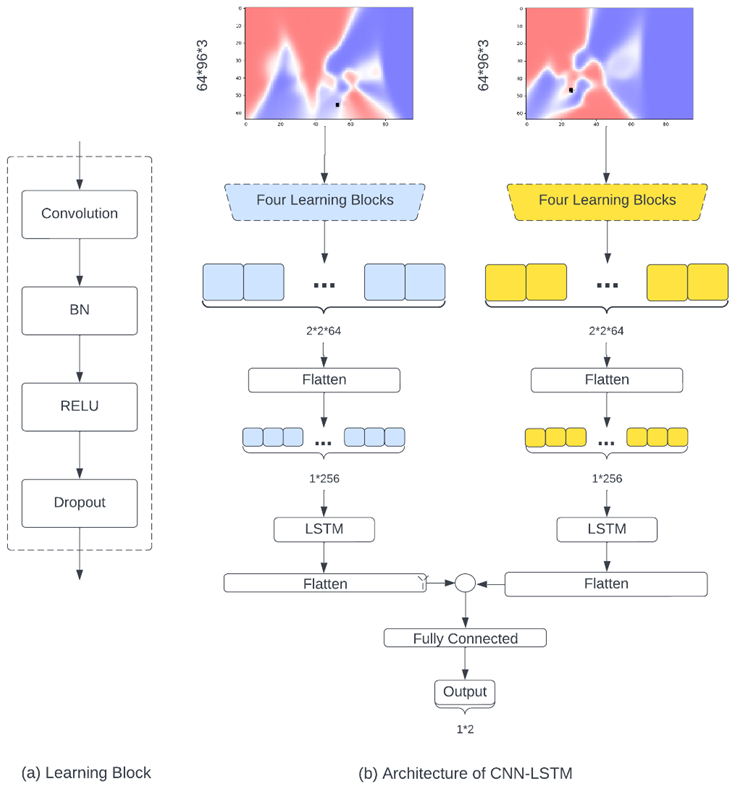}
    \caption{(a) local feature learning block. Batch Normalization and Rectified Linear Unit are abbreviated as BN and ReLU. (b) CNN-LSTM Classifier architecture}
    \label{fig:CNN-LSTM}
\end{figure}

To build the Pitch Control Map classifier, we utilize CNN-LSTM architecture containing Convolutional Neural Network (CNN) and Long Short-Term Memory Network (LSTM). The architecture is shown in Figure \ref{fig:CNN-LSTM}. 
We first construct the encoder, a group of neural networks with Convolutional Layers and Fully-Connected Layers embedded to encode Pitch Control Maps at t and t-1 to low-dimensional vector. Then we downsample the two vectors, including extract information from input data,  to further lower the dimensions. These vectors go through the LSTM afterwards. The output vectors of LSTM are concatenated to one vector and it is downsampled by a dense layer to have the same length as input. The classifier is trained by minimizing the Cross Entropy Loss between real label and predicted label vector.
There are two reasons for using part of the dataset to pre-train a classifier rather than manual labelling the whole dataset. Firstly, the whole dataset is too large to do precise manual labelling. Secondly, the sequences showing ambiguous reliance between two frames can be labelled with higher validity by trained classifier than manual work.

\subsection{Conditional VRNN Model}
\begin{figure}
    \centering
    \includegraphics[width=\linewidth]{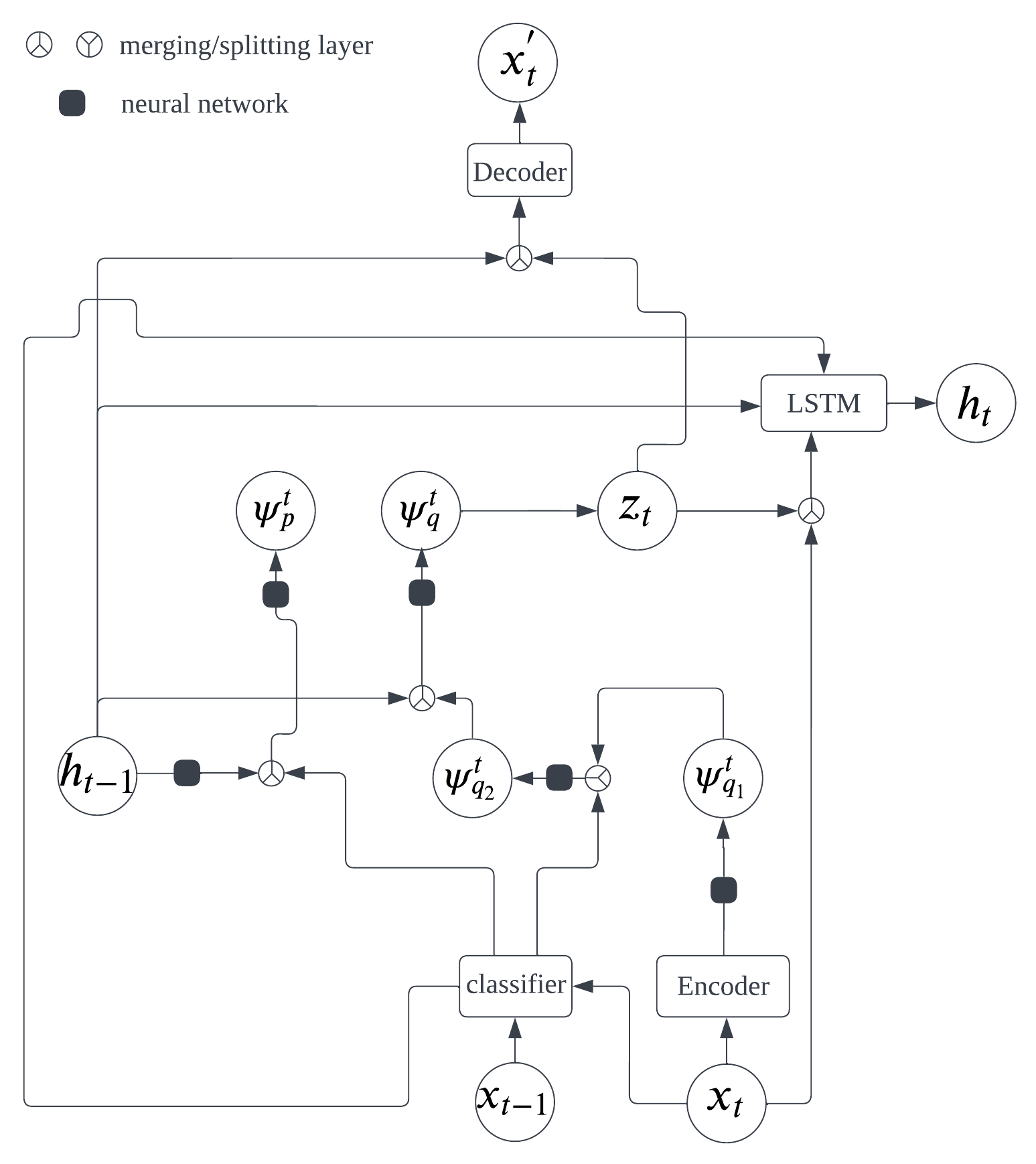}
    \caption{Architecture of Our Conditional VRNN Model}
    \label{fig:CVRNN}
\end{figure}

To imitate the team performance in different football contexts and build the performance benchmark, we propose a Conditional VRNN model to model the team movements and interactions in attacking and defending in terms of space control of the pitch.
A Vanilla VRNN model is able to generate new sequences through random sampling the latent space. The random sampling process make the generated samples have weak connections with contextual information in new environments. Specifically, in a new football context, when given only the beginning game state, Vanilla can hardly make stable long-term predictions due to the lack of conditioning on the contextual factors such as the changing pattern within each two adjacent frames. In our proposed model, we condition the reference, generation and temporal connections between latent spaces of each two adjacent input frames on the frame labels generated by pre-trained classifier. The architecture of model is shown in Figure \ref{fig:CVRNN}, in which \(x_t\) represents the input frame at time \(t\), \(h_t\) represents the hidden state of LSTM at time \(t\), and \(z_t\) denotes the latent space vector at time \(t\).

\subsubsection{Conditional Variable}
Within a given sequence, each frame except the first is labeled as pushing, backing or staying by utilizing the pre-trained classifier. This label shows the connection between the current frame and the previous one in terms of the change in controlled area of the pitch. We generate the conditional variable for each concerned frame by one-hot encoding the label. We condition the prior, posterior and LSTM parameters on the conditional variable when training the model.

\subsubsection{Frame Encoder}
The encoder consists of strided convolutional layers and Rectified Linear Units. It extracts latent representations from each frame from input sequences.
\subsubsection{Prior and Approximate Posterior}
\(\psi_p^t\), the prior at time \(t\), is parameterized as a diagonal normal distribution \(\mathcal{N}(\mu_{0,t},diag(\sigma_{0,t}^2))\) in which \(\mu_{0,t}\) and \(\sigma_{0,t}\) are conditioned on \(h_{t-1}\), the hidden state vector of LSTM and \(a_t\), the conditional variables. The prior distribution follows the distribution:
\begin{equation}
    \psi_p^t=\left[\mu_{0,t},\sigma_{0,t}\right]=\psi_\tau(\psi_\tau\left(h_{\left(t-1\right)}\right),a_t)
\end{equation}
\(\psi_q^t \), the approximate posterior at time \(t\), is parameterized as another diagonal normal distribution \(\mathcal{N}\left(\mu_{z,t},diag\left(\sigma_{z,t}^2\right)\right)\) in which \(\mu_{z,t}\) and \(\sigma_{z,t}\) are conditioned on the encoded input frame, \(a_t\), the conditional variables and \(h_t-1\), the hidden state vector at \(t-1\). The posterior distribution is shown as follows.
\begin{equation}
    \begin{split}
        & \psi_q^t=\left[\mu_{z,t},\sigma_{z,t}\right]={\psi_\tau(\psi}_\tau(\psi_\tau\left(\psi_{enc}(x_t\right)),a_t),\psi_\tau(h_{t-1}))\ 
        \\
        & \psi_{q_1}^t= \psi_{enc}(x_t)
        \\
        & \psi_{q_2}^t=\psi_\tau(\psi_\tau\left(\psi_{enc}(x_t\right)),a_t)
    \end{split}
\end{equation}
\(\psi_\tau\) denotes the neural networks containing Fully-Connected and ReLU layers, and \(\psi_{enc}\) denotes the encoder.
\subsubsection{Frame Decoder}
At time \(t\), we concatenate \(z_t\) and \(\psi_\tau(h_{t-1})\), the extraction of hidden state vector, both of which having same dimensions, along the channel dimension. The concatenated vectors go through the decoder consisting of transposed convolutions and Rectified Linear Units.
\subsubsection{Recurrence}
To capture the temporal factors among input frames within a sequence, the LSTM is used to denote the recurrent connections among frames and help generate new sample sequences consistently. \(h_t\), the hidden state at time \(t\), is updated according to input \(x_t\), latent variables \(z_t\), conditional variables \(a_t\) and hidden state variables \(h_{t-1}\). The initial hidden state is set as a zero vector.
\subsubsection{Model Learning}
The loss function for training our model contains the KL divergence and reconstruction loss. The KL divergence is the difference between posterior and prior and the reconstruction loss is computed using SSIM (Structural Similarity Index Measure). We calculate the loss at each timestep of the sequence and the mean of losses across all timesteps is used as the loss of input sequence. The loss at time \(t\) is shown as follow: 
\begin{equation}
    \mathcal{L} = -KL(q(z_t|x_{\leq t}, z_{<t})\parallel p(z_t|x_{<t}, z_{<t}))+ L_{ssim}
\end{equation}

\section{Experiments}

\subsection{Evaluation Metric}
The data used for training Conditional VRNN model consists of Pitch Control Maps. To accurately evaluate the model performance, the structural difference rather than each two pixels' difference between two images is the key factor. If the structural patterns of blue and red areas from two Pitch Control Maps match with each other it can be affirmed that these two represent the same football game context \cite{spearman2017physics}, and the reconstructed image has good quality. 
Structural Similarity Index Measure (SSIM) is a perception-based model treating degradation as perceived change in structural information, used for measuring two images' difference \cite{wang2004image}. It incorporates luminance masking and contrast masking terms. SSIM function is composed by three parts, luminance \(l\), contrast \(c\), and structure \(s\) with three positive weights, \(\alpha\), \(\beta\), and \(\gamma\).
\begin{equation}
    SSIM_{(x,y)}=[l(x,y)]^\alpha*[c(x,y)]^\beta*[s(x,y)]^\gamma
\end{equation}
In the work from Flynn et al, SSIM is proved to be a objective image quality metric \cite{flynn2013image}. They invite human participants to perceive the difference between images with different SSIM difference. As shown in the results, when SSIM is over 0.95, a human cannot detect the difference between two images.
Compared with other measuring metrics such as Cross Entropy and Mean Squared Error, SSIM estimates the structural differences rather than absolute differences \cite{kline2005revisiting,sara2019image}. Therefore, we use SSIM instead of Cross Entropy for evaluating the model performance.
\subsection{Quantitative Results}
We apply the SSIM measurement to evaluate the performance of Conditional VRNN model (Our CVRNN) on testing dataset. We run the same experiment with three other models for comparison. These three model includes, CVRNN 1, the VRNN model with LSTM parameters conditioned on label vectors, CVRNN 2, the VRNN model with prior and posterior conditioned on label vectors and a Vanilla VRNN model. For each test sequence, mean SSIM is calculated across all the frames. Then it is averaged across all the sequences in testing dataset.
Firstly, we evaluate the model performance with all frames included for each input sequence. Secondly, we evaluate the model performance with only the first frame for each sequence as input. The results are shown in Table \ref{tab:SSIM1} and Table \ref{tab:SSIM2} respectively.

\begin{table}[H]
    \centering
    \resizebox{8.5cm}{!}{
        \begin{tabular}{lrrrr}
            \toprule
               &Our CVRNN  & CVRNN 1 & CVRNN 2  & Vanilla VRNN \\
            \midrule
            SSIM     &0.954 $\pm 0.04$ \%     & 0.950 $\pm 0.06$ \%    & 0.952 $\pm 0.07$ \%    & 0.9 $\pm 0.10$ \% \\
            \bottomrule
        \end{tabular}
        }
    \caption{Evaluation of four models in sequence reconstruction using SSIM measurement}
    \label{tab:SSIM1}
\end{table}
\begin{table}[H]
    \centering
    \resizebox{8.5cm}{!}{
        \begin{tabular}{lrrrr}
            \toprule
               &Our CVRNN  & CVRNN 1 & CVRNN 2  & Vanilla VRNN \\
            \midrule
            SSIM     &0.850 $\pm 0.12$ \%     & 0.846 $\pm 0.15$ \%    & 0.847 $\pm 0.16$ \%    & 0.795 $\pm 0.25$ \% \\
            \bottomrule
        \end{tabular}
        }
    \caption{Evaluation of four models in sequence prediction using SSIM measurement}
    \label{tab:SSIM2}
\end{table}
From Table \ref{tab:SSIM1}, our model outperforms the others with higher SSIM, which indicates that the sequence reconstructed by our model is closer to the input sequence in terms of structural similarity.
From Table \ref{tab:SSIM2}, our model has better performance in terms of imitating the patterns of ground truth input compared to the other three when making long-term generation given only the first frame.
\subsection{Qualitative Results}
\begin{figure}[H]
    \centering
    \includegraphics[width=3in]{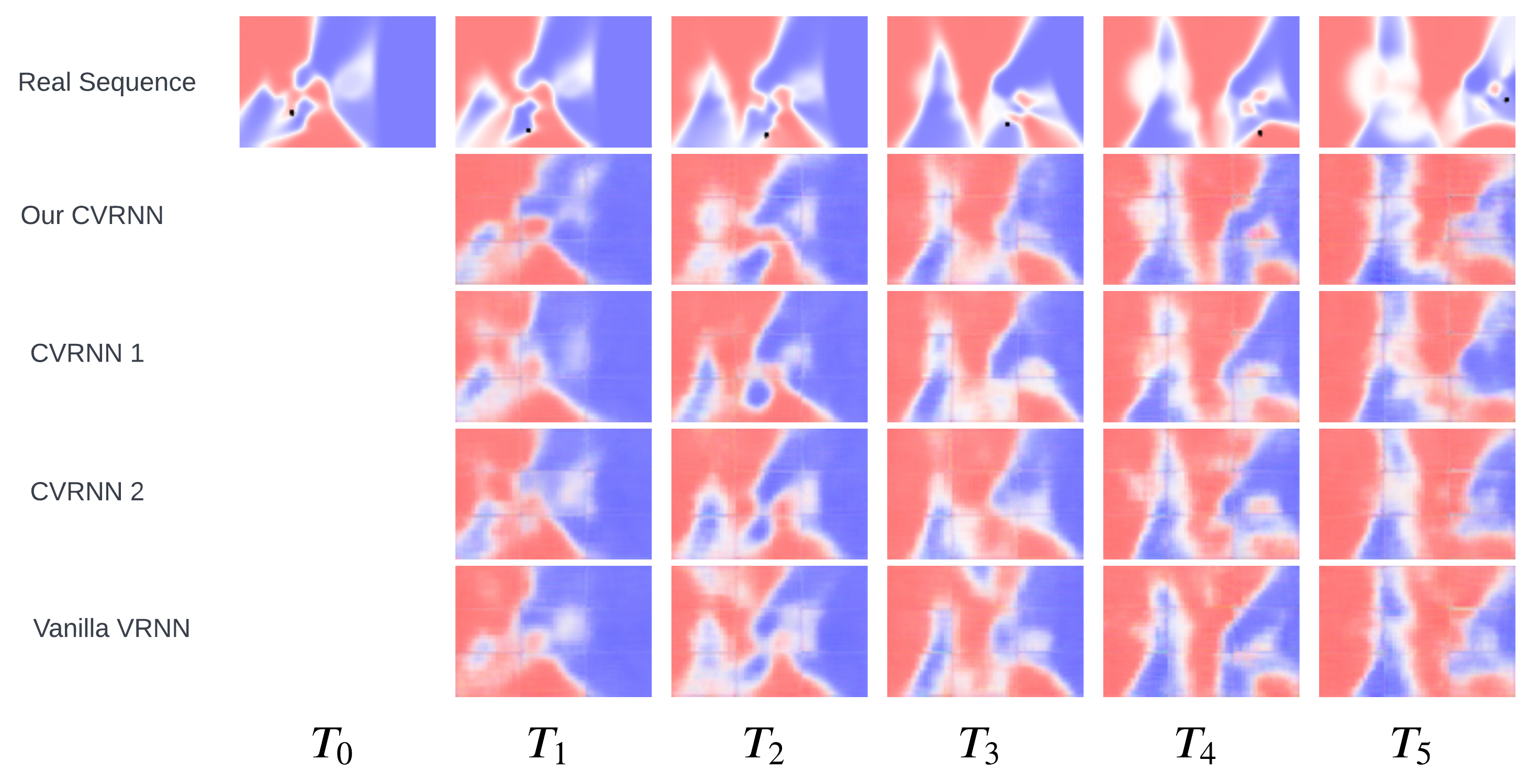}
    \caption{Pitch Control Map sequence reconstruction}
    \label{fig:rec}
\end{figure}

For a given possession of game-play, we compare the sequences reconstructed by our model and the others. The results are shown in Figure \ref{fig:rec}. All four groups of sequences match the input sequence well with great image quality at each timestep. It suggests that our model and the others perform well in capturing the contextual factors from input and reconstructing it.

\begin{figure}[H]
    \centering
    \includegraphics[width=3.3in]{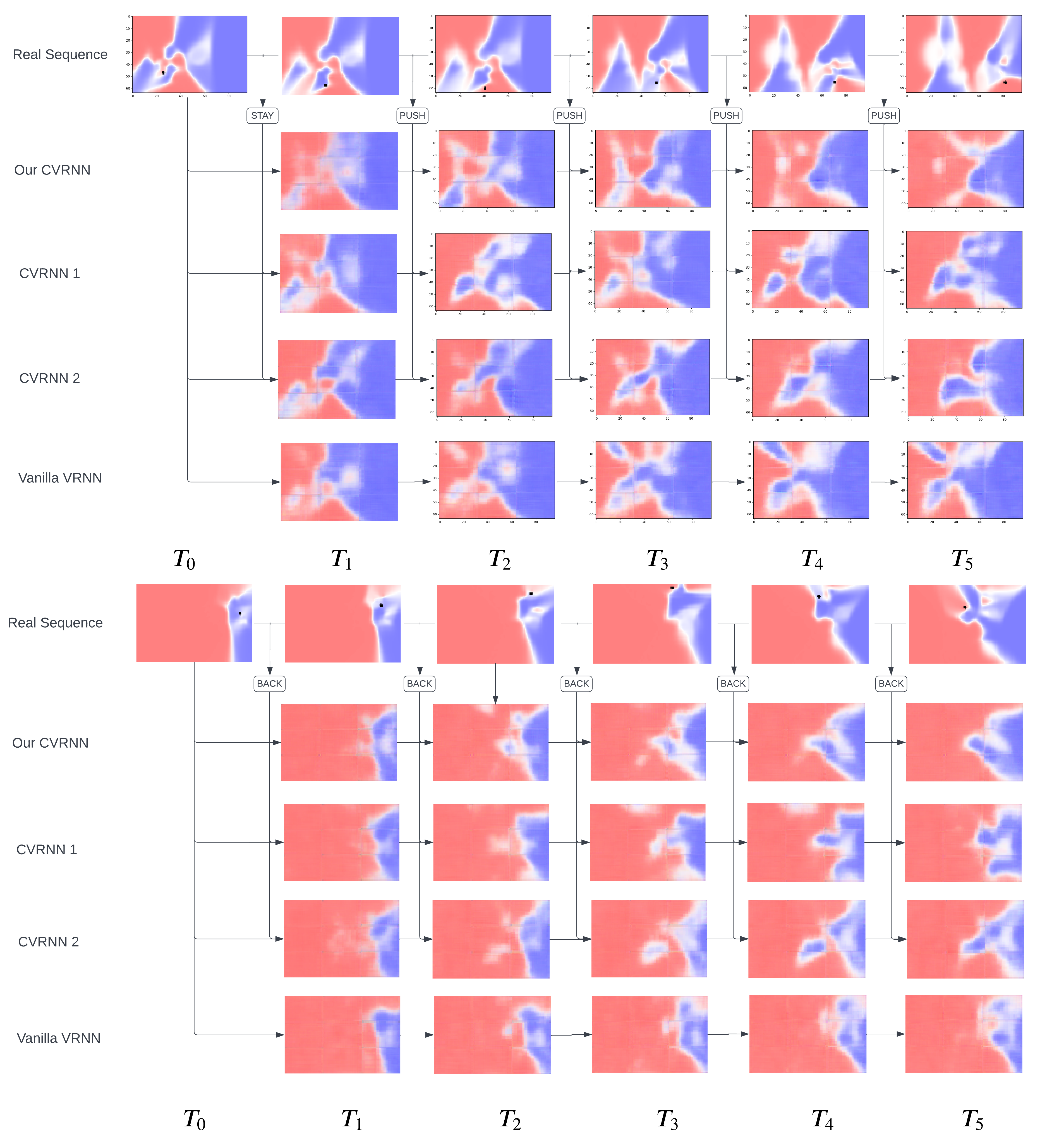}
    \caption{Pitch Control Map sequence predictions in two different possessions}
    \label{fig:pred}
\end{figure}

Then we apply our model on a sequence of game play with only first frame and label vectors for each frame after the first one to predict the sequence. The prediction made by our model and the other three in two different possessions are shown in Figure \ref{fig:pred}. Among all the models, Vanilla VRNN has the worst performance in terms of predict the changing patterns from real sequence. Moreover, the prediction of our model has a closer match to real sequence in both pushing and backing scenarios, compared to CVRNN 1 and CVRNN 2. It indicates our model is more capable of imitating real team performance than the others. More results can be checked in appendix.

\subsection{Computational Benchmark for Evaluating Human Performance}
With the trained Conditional VRNN model, we can imitate the team performance in given contexts of football game. We test the effectiveness of using this model as benchmark versus using the existing metric to evaluate team performance. To enable the comparison, we quantify the value of Pitch Control Maps by utilizing EPV \cite{rudd_2011}. 

\begin{figure}
    \centering
    \includegraphics{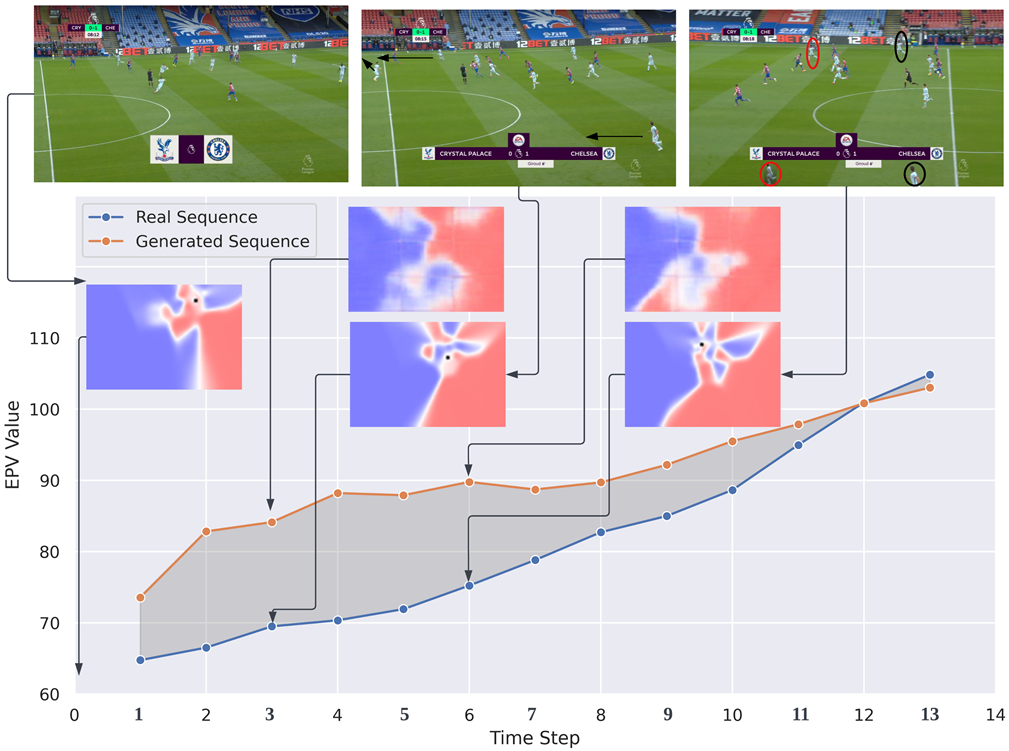}
    \caption{Evaluating real sequence according to sequence generated by benchmark model (red and blue area denote the space controlled by attacking and defensive sides respectively)}
    \label{fig:benchmark}
\end{figure}

As shown in Figure \ref{fig:benchmark}, the example possession for test consists of 14 frames in which we use the first frame as the input to generate the benchmark sequence and compare the real sequence against this benchmark using EPV. The blue line shows the team performance according to the existing metric, EPV across the 14 timesteps while the orange line denotes the benchmark performance which the team is expected to reach in given context. Only using existing metric EPV, it can be inferred that the attacking team is doing good with consistently increasing high EPV values and defensive side is vulnerable letting opponent take control of lots of space. However, in real case, the defensive side is quite efficient in preventing the opponent from converting a counter-attack to a scoring chance while attacking side fails to convert the counter-attack to real threat. At the beginning of this possession, attacking team wins the ball and is ready to make a counter-attack given that the opponent defenders are out of position for attacking leaving lots of open space behind \cite{sarmento2014patterns,armatas2005analysis}. According to our benchmark model, the team in this type of context will utilize the open space from left and right wings and quickly push their formation towards the opponent's half as shown in the generated Pitch Control Maps at timestep 3 and timestep 6. However, the attacking side fails to do so due to two factors. Firstly, according to the real game scenes at timestep 3 and timestep 6, the defensive side makes a quick pressure after losing control of the ball and moves back as a whole team instantly after the opponent player drives past the halfway line. This lets them keep the formation compact, reducing the opponent's threat from counter-attack \cite{sarmento2013regular}. Secondly, from the Pitch Control Maps and real game scenes at timestep 3 and timestep 6, it can be argued that the attacking team wingers haven't reacted quickly enough in terms of making forward runs when the team wins the ball, which causes their team to miss the short window for making successful counter-attack \cite{hughes2019transition}. In this scenario, by using our benchmark, we can tell that defending team outperforms the benchmark model prediction and attacking side falls behind the expectation, which matches what has happened in real life while the existing metric makes inaccurate indications. In general, by imitating the real-life human performance in complicated contexts through multi-agent generative modelling and providing, we manage to capture key contextual factors when making human performance evaluation.
\section{Conclusion}
\begin{figure}
    \centering
    \includegraphics[width=2.5in]{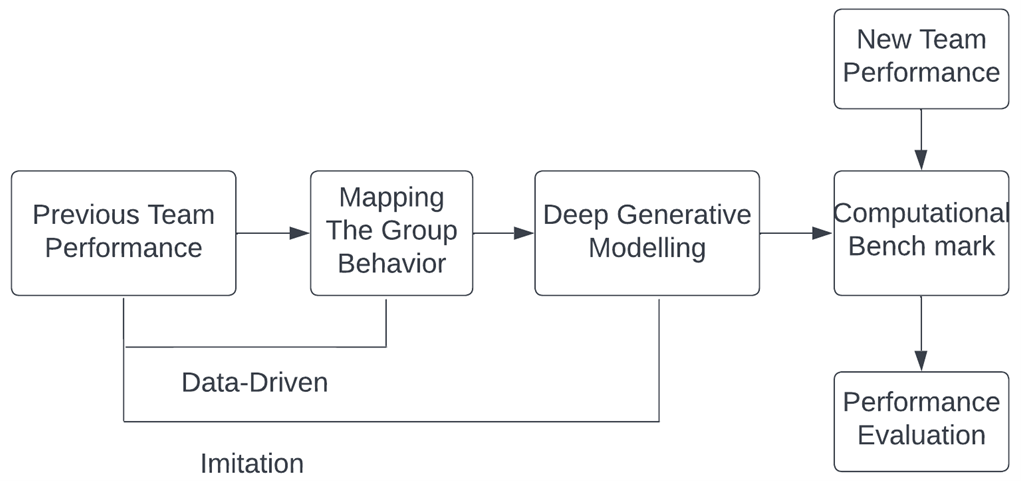}
    \caption{Framework for human performance evaluation}
    \label{fig:overall}
\end{figure}

In this work, we propose a deep generative model to learn the behavior patterns and interactions among agents in multi-agent system. We do the research in the context of professional football by mapping the previous human performance in terms of tracking and event data to Pitch Control Maps and training a deep generative model on the mapped human performance to learn the player moving patterns and interactions within football games and imitate the team performance in new given context. This imitated team performance is used as a benchmark for evaluating real team performance. As shown in the results on a Premier league dataset, our model is a reliable computational benchmark for evaluating human performance in football. It is believed to be a useful addition to existing state-of-the-art performance evaluation metric in football. Moreover, the proposed method to evaluate human performance in football can be applied in other areas in which there is a team of cooperating humans participating to complete certain tasks. The overall framework can be summarized as Figure \ref{fig:overall}. The biggest challenge of applying this framework is to map team behaviors to a demonstration which can be sampled from for training the generative model. Nonetheless, considering the efficiency of applying our model in evaluating human performance in football which is a extremely complex environment containing a lot of contextual factors, the model with similar structure can be a promising addition to existing evaluation metrics of other areas.

\appendix

\section{Visualization of Reconstruction Results of Our CVRNN Model}

\begin{figure}[H]
    \centering
    \includegraphics[width=\linewidth]{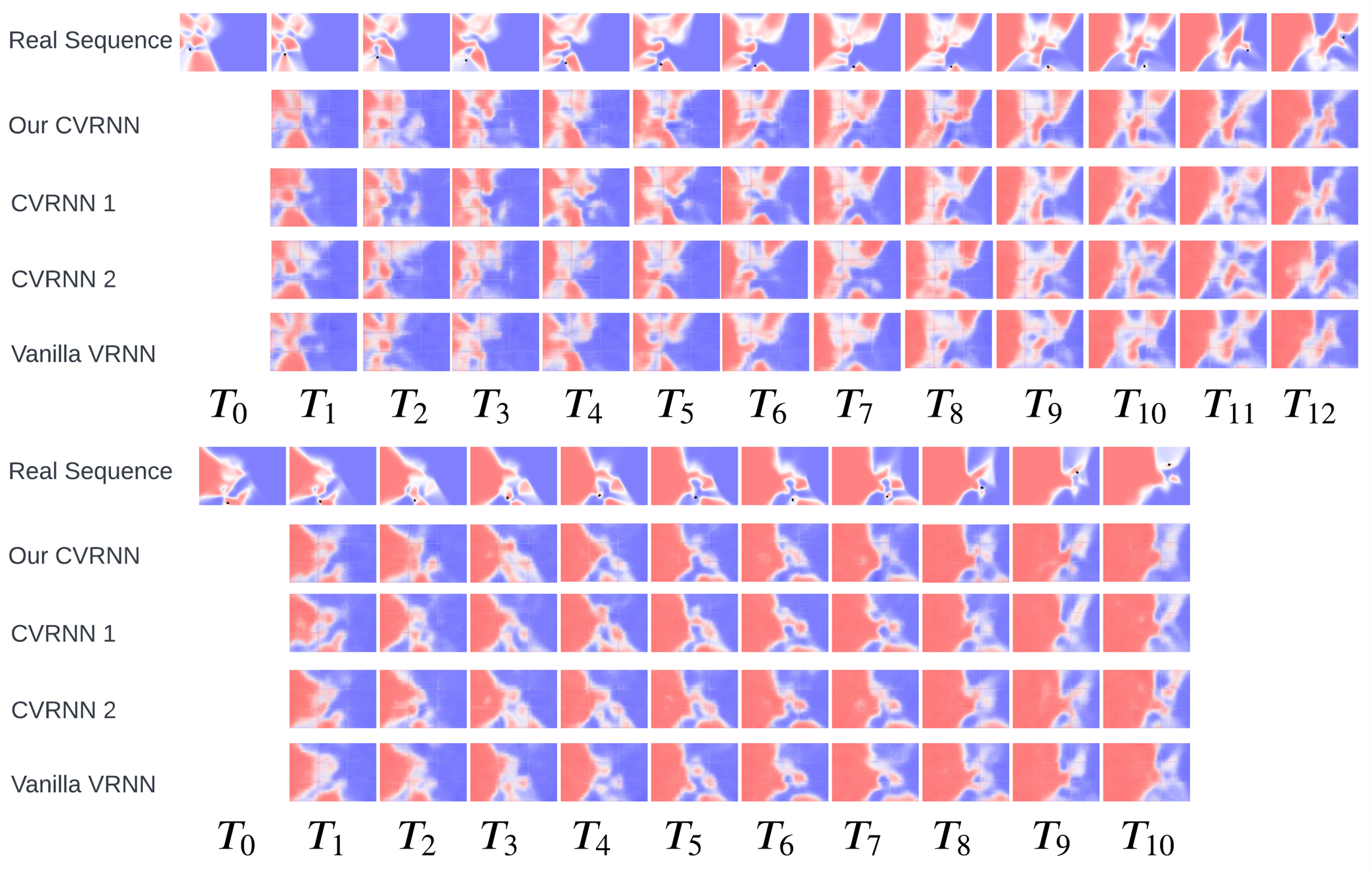}
    \caption{Comparison of model performance in sequence reconstruction from two different possessions}
    \label{fig:Appendix_a}
\end{figure}

\section{Visualization of Prediction Results of Our CVRNN Model}

\begin{figure}[H]
    \centering
    \includegraphics[width=\linewidth]{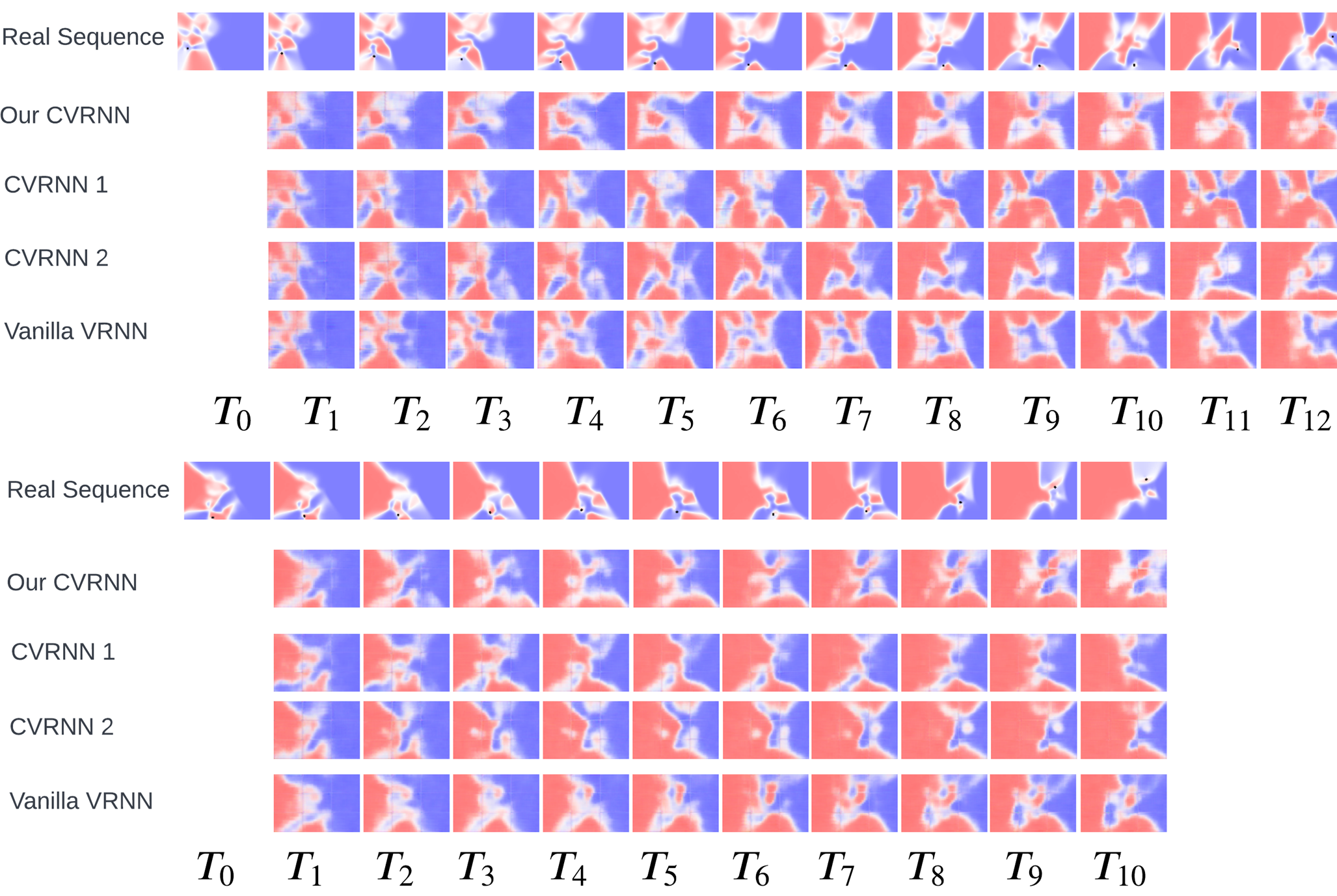}
    \caption{Comparison of model performance in sequence Prediction from two different possessions}
    \label{fig:Appendix_b}
\end{figure}

\bibliographystyle{named}
\bibliography{ijcai23}

\end{document}